\documentclass[review]{elsarticle}
\usepackage{hyperref}

\usepackage{epsfig}
\usepackage{graphicx}
\usepackage{amsmath}
\usepackage{algorithmic}
\usepackage[ruled]{algorithm2e}

\journal{Pattern Recognition}

\begin{document}
\begin{frontmatter}

\title{Deep Feature Learning with Relative Distance Comparison for Person Re-identification}

\author{\begin{small}Shengyong Ding , Liang Lin , Guangrun Wang , Hongyang Chao \end{small}}
\address{Sun Yat-sen University, Guangzhou 510006, China}





\begin{abstract}
Identifying the same individual across different scenes is an important yet difficult task in intelligent video surveillance. Its main difficulty lies in how to preserve similarity of the same person against large appearance and structure variation while discriminating different individuals. In this paper, we present a scalable distance driven   feature learning framework based on the deep neural network for person re-identification, and demonstrate its effectiveness to handle the existing challenges. Specifically, given the training images with the class labels (person IDs), we first produce a large number of triplet units, each of which contains three images, i.e. one person with a matched reference and a mismatched reference. Treating the units as the input, we build the convolutional neural network to generate the layered representations, and follow with the $L2$ distance metric. By means of parameter optimization, our framework tends to maximize the relative distance between the matched pair and the mismatched pair for each triplet unit. Moreover, a nontrivial issue arising with the framework is that the triplet organization cubically enlarges the number of training triplets, as one image can be involved into several triplet units. To overcome this problem, we develop an effective triplet generation scheme and an optimized gradient descent algorithm, making the computational load mainly depends on the number of original images instead of the number of triplets. On several challenging databases, our approach achieves very promising results and outperforms other state-of-the-art approaches.
\end{abstract}

\begin{keyword}
Person Re-identification, Deep Learning, Distance Comparison
\end{keyword}

\end{frontmatter}


\section{Introduction}
\begin{figure}[!htb]
 \begin{center}
\includegraphics [width=10cm]{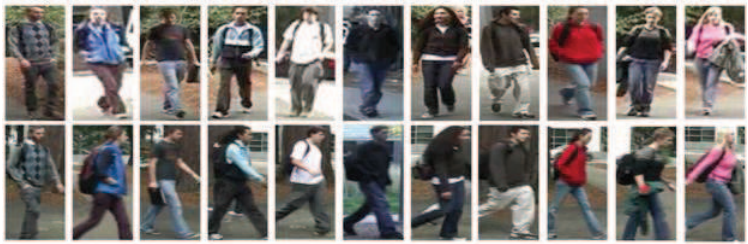}
 \caption{Typical examples of pedestrians shot by different cameras. Each column corresponds to one person. Huge variations exist due to the light, pose and view point changes.}
 \label{fig:challenges}
 \end{center}
\end{figure}

Person re-identification, the aim of which is to match the same individual across multiple cameras, has attracted widespread attention in recent years due to its wide applications in video surveillance. It is the foundation of threat detection, behavioral understanding and other applications. Despite the considerable efforts of computer vision researchers, however, it is still an unsolved problem due to the dramatic variations caused by light, viewpoint and pose changes \cite{lin2009stochastic}. Figure $\ref{fig:challenges}$ shows some typical examples from two cameras.

There are two crucial components, i.e. feature representations and distance metric in person re-identification systems. In these two components, feature representation is more fundamental because it is the foundation of distance learning. The features used in person re-identification range from the color histogram \cite{gray2008viewpoint}, spatial cooccurrence representation model \cite{wang2007shape}, attributes model \cite{layne2012towards} to combination of multiple features \cite{gray2008viewpoint,farenzena2010person}. These handcrafted features can hardly be optimal in practice because of the different viewing conditions that prevail \cite{lin2015adaptive}. Given a particular feature representation, a distance function is learned to construct a similarity measure \cite{li2013human,li2013learning} with good similarity constraints . Although the effectiveness of the distance function has been demonstrated, it heavily relies on the quality of the features selected, and such selection requires deep domain knowledge and expertise \cite{lin2009stochastic}.

\begin{figure*}[!htb]
\begin{center}
\includegraphics [width=11cm]{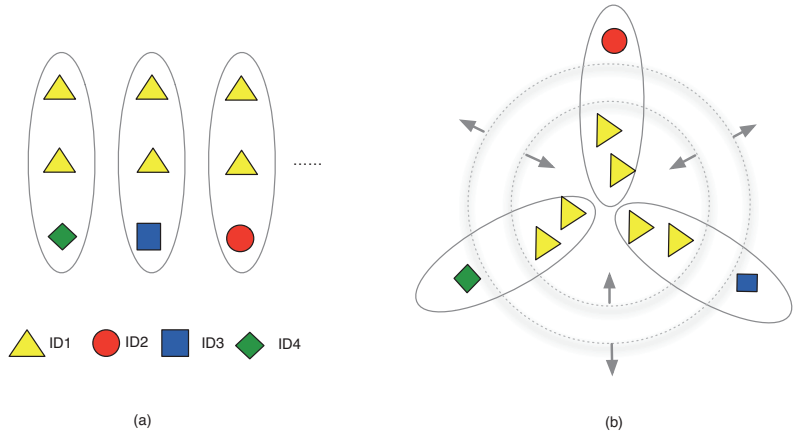}
\caption{Illustration of deep feature learning via relative distance maximization. The network is trained by a set of triplets to produce effective feature representations with which the true matched images are closer than the  mismatched images. }
\label{fig:framework}
\end{center}
\end{figure*}

In this paper, we present a scalable distance driven feature leaning framework via the convolutional network to learn representations for the person re-identification problem. Unlike the traditional deep feature learning methods aimed at minimizing the classification error, in our framework, features are learned to maximize the relative distance. More specifically, we train the network through a set of triplets. Each triplet contains three images, i.e. a query image, one matched reference (an image of the same person as that in the query image) and one mismatched reference. The network produces features with which the $L_2$ distance between the matched pair and the mismatched pair should be as large as possible for each triplet. This encourages the distances between matched pairs to take smaller values than those between the mismatched pairs. Figure $\ref{fig:framework}$ illustrates the overall principles. As discussed in \cite{zheng2011person}, the triplet-based model is a natural model for the person re-identification problem for two main reasons. First, the intra-class and inter-class variation can vary significantly for different classes, and it may thus be inappropriate to require the distance between a matched pair or mismatched pair to fall within an absolute range. Second, person re-identification training images are relatively scarce, and the triplet-based training model can generate more constraints for distance learning, thereby helping to alleviate the over-fitting problem.

Similar to traditional neural networks, our triplet-based model also uses gradient descent algorithms in solving the parameters. Owing to limitations in memory size, it is impossible to load all the triplets for a given labeled image set into the memory to calculate the gradient. A practical means is to train the network iteratively in mini-batches, that is, in each iteration, a subset of the triplets are generated and the network parameters are then updated with the gradient derived from that batch. However, as we will see in the later sections, randomly generating the triplets at each iteration is inefficient as only a small number of distance constraints are imposed on the images within the triplets. Therefore we propose a more efficient triplet generation scheme. In each iteration, we randomly select a small number of classes (persons) from the dataset and generate the triplets using only those images, which guarantees that only a small number of images are selected in each iteration and rich distance constraints are imposed. In our proposed triplet generation scheme, one image can occur in several triplets in each iteration with a high degree of probability, and we thus design an extended network propagation algorithm to avoid recalculating the gradients of the same images. Our triplet generation scheme and the extended network propagation algorithm render the overall computational load of our model dependent mainly on the number of the training images, not on the number of triplets. Our approach also enables us to use the existing deep learning implementations to solve our model parameters with only slight modifications. In summary, we make two contributions to the literature:

1) A scalable deep feature learning method for person re-identification via maximum relative distance.

2) An effective learning algorithm for which the training cost mainly depends on the number of images rather than the number of triplets.

The remainder of this paper is organized as follows. In section two, we review the related work on person re-identification problems. In section three, we present our formulation and network architecture. In section four, we derive the algorithms for solving the model parameters using gradient descent methods for a small triplet set.  In section five, we show how to train the network in batch mode with an efficient triplet generation scheme, and in section six, we present our experimental results. Section seven concludes our work.

\section{Related Work}

Feature representation and distance metric are the two main components of person re-identification systems. The existing approaches to person re-identification tasks primarily make use of handcrafted features such as color and texture histograms \cite{gray2008viewpoint, wang2007shape, lin2012representing}. To increase their representative capability, features have been designed to carry spatial information \cite{gray2008viewpoint, farenzena2010person}. For example, Farezena et al. utilized the symmetry property of pedestrian images to propose a method called Symmetry Driven Accumulation of Local Features (SDALF) which is robust to background clutter \cite{farenzena2010person}.  The body configuration-based pictorial structure features have been also well studied to cope with individual variations \cite{lin2015AOGraph,xu2013human}. 

In addition to handcrafted feature designs, some studies addressed learning features for person re-identification tasks. For example, Gray and Tao \cite{gray2008viewpoint} proposed the use of Adaboost to learn effective representations from an ensemble of local features. Zhao et al.  \cite{zhao2013learning} proposed the learning of mid-level features from hierarchical clusters of patches.

Another important research direction in person re-identification is distance learning. Zheng et al. \cite{zheng2011person} formulated the distance learning as a Probabilistic Relative Distance Comparison model (PRDC) to maximize the likelihood that correctly matched pairs will have a smaller distance between them than incorrectly matched pairs. In addition, Mignon and Jurie proposed Pairwise Constrained Component Analysis (PCCA) to project the original data into a lower dimensional space \cite{mignon2012pcca}, in which the distance between pairs has the desired properties. Li et al. introduced a locally adaptive thresholding rule to metric learning models (LADF), and reported that it achieved good performance on person re-identification tasks \cite{li2013learning}. RankSVM has also  been proposed for learning a subspace in which the matched images have a higher rank than the mismatched images for a given query. There are also a number of general distance learning methods that have been rarely exploited in the context of person re-identification problems \cite{xing2002distance, weinberger2005distance, davis2007information, xiang2008learning}.

Inspired by the success of deep learning, there are also some literatures applying neural network models to address the person re-identification problems. Dong Yi et al. \cite{DBLP:journals/corr/YiLL14} applied a deep neural network to learn pair-wise similarity and achieved state-of-the-art performance. Hao Liu et al. \cite{Liu20151283} presented a Set-Label Model, which applies DBN (Deep Belief Network) and NCA (Neighborhood Component Analysis) on the proposed concatenated features of the query image and the gallery image to improve the person re-identification performance. Xu et al. \cite{xu2013human} adopted a cluster sampling algorithm \cite{lin2010layered} for re-identifying persons with templates. Li et al. \cite{li2013deepreid} proposed a deep learning framework for learning filter pairs that tries to automatically encode the photometric transforms across cameras. Our work differs from these methods in its loss function and learning algorithm.

The model most similar to that proposed herein was introduced by Wang et al. \cite{wang2014learning} and involved learning features for fine-grained image retrieval. They borrowed the network architecture designed by Krizhevsky et al. \cite{krizhevsky2012imagenet}, and pre-trained the network using soft-max loss function. It is unclear whether the triplet-based deep model can be effectively trained from triplets without other pre-training techniques. Here,  we extend the triplet-based model to the person re-identification problem with an efficient learning algorithm and triplet generation scheme. We demonstrate its effectiveness without pre-training techniques using a relatively simple network .

\section{Model}

\subsection{Formulation}
\begin{figure*}[!htb]
\begin{center}
\includegraphics [width=10cm]{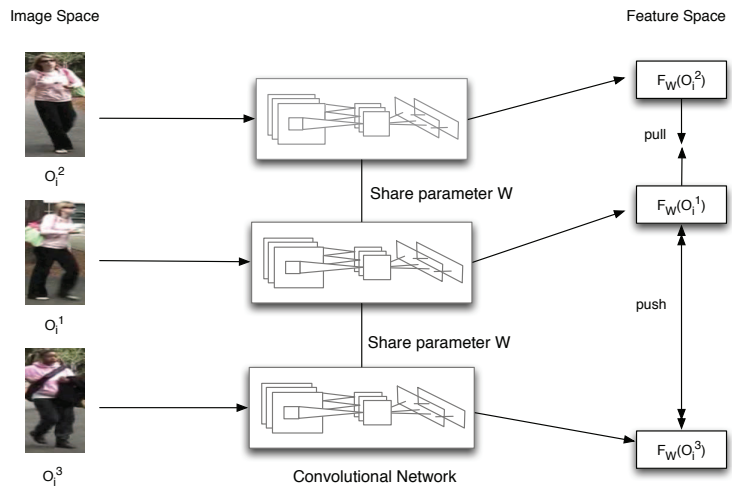}
\caption{Illustration of maximizing the distance for person re-identification. The $L_2$ distance in the feature space between the matched pair should be smaller than the  mismatched pair in each triplet.}
\label{fig:triplet-model}
\end{center}
\end{figure*}

Our objective is to use a deep convolutional network to learn effective feature representations that can satisfy the relative distance relationship under the $L_2$ distance. In other words, we apply a deep convolutional network to produce the feature for each image. And with these generated features, we require the distances between matched pairs should be smaller than those between mismatched pairs as depicted in Figure $\ref{fig:triplet-model}$. 

In our model, the relative distance relationship is reflected by a set of  triplet units $\{O_i\}$ where $O_i=<O_i^1,O_i^2,O_i^3>$, in which $O_i^1$ and $O_i^2$ are a matched pair and $O_i^1$ and $O_i^3$ are a mismatched pair. Let $W=\{W_j\}$  denote the network parameters and $F_W(I)$  denote the network output of image $I$, i.e. feature representation for image $I$. For a training triplet $O_i$, the desired feature should satisfy the following condition under  the $L_2$ norm.
\begin{small}
\begin{equation}
||F_W(O_i^1)-F_W(O_i^2)||<||F_W(O_i^1)-F_W(O_i^3)||
\end{equation}
\end{small}
or equally:
\begin{small}
\begin{equation}
||F_W(O_i^1)-F_W(O_i^2)||^2<||F_W(O_i^1)-F_W(O_i^3)||^2
\end{equation}
\end{small}

Here, we use the squared form to facilitate the partial derivative calculation. For a given training set $O$=$\{O_i\}$, the relative distance constraints are converted to the minimization problem of the following objective, i.e. maximizing the distance between matched pairs and mismatched pairs, where $n$ is the number of the training triplets.
\begin{small}
\begin{equation}
\label{equ:objFunc}
f(W,O)=\\
\Sigma_{i=1}^{n}\max\{||F_W(O_i^1)-F_W(O_i^2)||^2-||F_W(O_i^1)-F_W(O_i^3)||^2,C\}
\end{equation}
\end{small}
 The role of the max operation with the constant $C$ is to prevent the overall value of the objective function from being dominated by easily identifiable triplets, which is similar to the technique widely used in hinge-loss functions. We set $C$=-1 throughout the paper.

Note the network in our model still takes one image as input both for training and testing as the conventional convolutional network does. The triplet-based loss function is introduced for parameter optimization in the training stage. During the testing, we feed each testing image to the trained network to get its feature and use these features for performance evaluation under the normal $L_2$ distance.

\subsection{Network Architecture}
\begin{figure*}[!htb]
\begin{center}
\includegraphics [width=12cm]{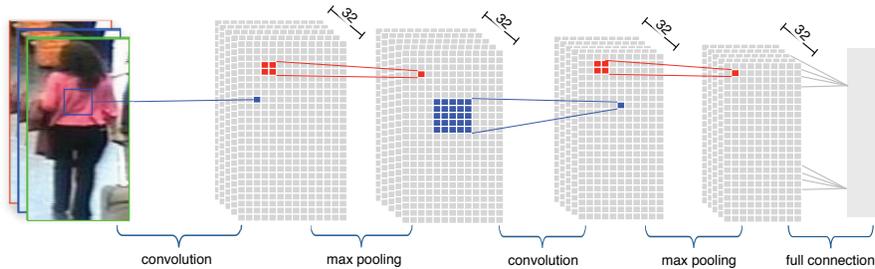}
\caption{An illustration of the network architecture. The first and third layers are convolutional layers and the second and fourth layers are max pooling layers. The final layer is a full connection layer.}
\label{fig:network}
\end{center}
\end{figure*}

All existing person re-identification datasets are relatively small, and we thus designed a simplified network architecture for our model. Figure $\ref{fig:network}$ shows the overall network architecture, which comprises five layers. The first and third layers are convolutional layers and the second and fourth layers are pooling layers. The first convolutional layer includes 32 kernels of size 5$\times$5$\times$3 with a stride of 2 pixels. The second convolutional layer takes the pooled output of the first convolutional layer as input and filters it with 32 kernels of size 5$\times$5$\times$32 with a stride of 1 pixel. The final 400 dimensional layer is fully connected to the pooled output of the second convolutional layer with the following normalization:

Let $\{x_i\}$ denote the output before normalization, with the normalized output then calculated by:
\begin{equation}
y_i=\frac{x_i}{\sqrt{\Sigma x_i^2}}
\end{equation}

Note that this normalization differs from the normalization scheme applied by Krizhevsky et al. \cite{krizhevsky2012imagenet} over different channels. Our normalization ensures that the distance derived from each triplet cannot easily exceeds the margin $C$ so that more triplet constraints can take effect for the whole objective function. Accordingly, the back propagation process accounts for the normalization operation using the chain rule during calculation of the partial derivative.

We use overlapped max pooling for the pooling operation. More precisely, the pooling operation can be thought of as comprising a grid of pooling units spaced $s$ pixels apart, with each summarizing a neighborhood of size $z\times z$ centered at the location of the pooling unit. We set $s$=1 and $z$=2 in our network. For the neuron activation functions, we use Rectified Linear Units to transform the neuron inputs, thereby speeding up the learning process and achieving good performance, as discussed in \cite{krizhevsky2012imagenet}.

\section{Learning Algorithm}
In this section, we show how to solve the network given a fixed set of training triplets. We assume the memory is sufficiently large to load all of the triplets. The procedures for generating triplets from labeled images and training the network using the batch mode is relegated to the next section.
\subsection{Triplet-based gradient descent algorithm}
We first present a direct learning algorithm derived from the definition of the objective function. For ease of exposition, we introduce $d(W,O_i)$, which denotes the difference in distance between the matched pair and the mismatched pair in the triplet $O_i$.
\begin{small}
\begin{equation}
d(W,O_i)=||F_W(O_i^1)-F_W(O_i^2)||^2-||F_W(O_i^1)-F_W(O_i^3)||^2
\end{equation}
\end{small}
and the objective function can be rewritten as,
\begin{small}
\begin{equation}
f(W,O)=\Sigma_{O_i}\max\{d(W,O_i),C\}
\end{equation}
\end{small}
Then the partial derivative of the objective becomes
\begin{small}
\begin{equation}
\label{equ:sum}
\frac {\partial f}{\partial W_j}=\Sigma_{O_i}  h(O_i)
\end{equation}
\end{small}
\begin{small}
\begin{equation}
\label{equ:max}
h(O_i)=\begin{cases}
\frac{\partial d(W,O_i)}{\partial W_j}, & \text{if } d(W,O_i)>C;\\
0 ,& \text{if } d(W,O_i)<=C;
\end{cases}
\end{equation}
\end{small}
By the definition of $d(W,O_i)$, we can obtain the gradient of  $d(W,O_i)$ as follows:
\begin{small}
\begin{equation}
\label{equ:partial}
\begin{split}
 \frac{\partial d(W,O_i)}{\partial W_j}=2(F_W(O_i^1)-F_W(O_i^2))'\cdot \frac{\partial{F_W(O_i^1)-\partial{F_W(O_i^2)}}}{\partial W_j}\\-2(F_W(O_i^1)-F_W(O_i^3))' \cdot \frac{\partial{F_W(O_i^1)-\partial{F_W(O_i^3)}}}{\partial W_j}
 \end{split}
\end{equation}
\end{small}
We can now see that the gradient on each triplet can be easily calculated given the values of $F_W(O_i^1),F_W(O_i^2),F_W(O_i^3)$ and $\frac{\partial F_W(O_i^1)}{\partial W_j},\frac{\partial F_W(O_i^2)}{\partial W_j},\frac{\partial F_W(O_i^3)}{\partial W_j}$, which can be obtained by separately running the standard forward and backward propagation for each image in the triplet. As the algorithm needs to go through all of the triplets to accumulate the gradients for each iteration, we call it the triplet-based gradient descent algorithm. Algorithm $\ref{alg:gradDescentByTriplet}$ shows the overall process.
\begin{algorithm}[htb]        
\caption{Triplet-based gradient descent algorithm }             
\label{alg:gradDescentByTriplet}                  
\begin{algorithmic}[1]                
\REQUIRE ~~\\                          
   Training samples $\{O_i\}$;
\ENSURE ~~\\                           
	The network parameters $\{W_j\}$
\WHILE    {$t<T$}
\STATE $t\leftarrow t+1$;
\STATE $\frac {\partial f}{\partial W_j}=0$

    \FORALL {training triplet $O_i$}
    \STATE  Calculate  $F_W(O_i^1)$,$F_W(O_i^2)$,$F_W(O_i^3)$ by forward propagation;
    \STATE Calculate $\frac{\partial{F_W(O_i^1)}}{\partial W_j}$，$\frac{\partial{F_W(O_i^2)}}{\partial W_j}$，$\frac{\partial{F_W(O_i^3)}}{\partial W_j}$ by back propagation;
    \STATE Calculate $\frac{\partial d(W,O_i)}{\partial W_j}$ according to equation $\ref{equ:partial}$;
    \STATE Increment the gradient $\frac {\partial f}{\partial W_j}$ according to equation $\ref{equ:sum}, \ref{equ:max}$;
       \ENDFOR
\STATE $W_j^t=W_j^{t-1}-\lambda_t \frac {\partial f}{\partial W_j}$;
\ENDWHILE          
\end{algorithmic}
\end{algorithm}
\subsection{Image-based gradient descent algorithm}

In the triplet-based gradient descent algorithm, the number of network propagations depends on the number of training triplets in each iteration, with each triplet involving three rounds of forward and backward propagation during the calculation of the gradient. However, if the same image occurs in different triplets, the forward and backward propagation of that image can be reused. Recognition of this potential shortcut inspired us to look for an optimized algorithm in which the network propagation executions depend only on the number of distinct images in the triplets. Before considering that algorithm, we first review the way in which  the standard propagation algorithm is deduced in the conventional CNN learning algorithm, where the objective function often takes the following form. Here $n$ is the number of training images.
\begin{equation}
f(I_1,I_2,...,I_n)=\frac{1}{n}\Sigma_{i=1}^{n}loss(F_W(I_i))
\label{equ:imLoss}
\end{equation}
As the objective function is defined as the sum of the loss function on each image $I_i$, we have:
\begin{equation}
\frac{\partial f}{\partial W}=\frac{1}{n}\Sigma_{i=1}^{n}\frac{\partial loss(F_W(I_i))}{\partial W}
\end{equation}
This shows that we can calculate the gradient of the loss function for each image separately and then sum these image-based gradients to obtain the overall gradient of the objective function. In the case of a single image, the gradient can be calculated recursively by the chain rule, which is given as follows.
\begin{equation}
\frac{\partial loss(F_W(I_i))}{\partial W^l}=\frac{\partial loss(F_W(I_i))}{\partial X_i^l} \frac{\partial X_i^{l}}{\partial W^l}
\label{equ:paramPartial}
\end{equation}
\begin{equation}
\frac{\partial loss(F_W(I_i))}{\partial X_i^l}=\frac{\partial loss(F_W(I_i))}{\partial X_i^{l+1}}\frac{\partial X_i^{l+1}}{\partial X_i^l}
\label{equ:feaPartial}
\end{equation}
In the above equations, $W^l$ represents the network parameters at the $l^{th}$ layer and $X_i^l$ represents the feature maps of the image $I_i$ at the same layer. The Equation $\ref{equ:paramPartial}$ holds because $X_i^l$ depends on the parameter $W^l$ and the Equation $\ref{equ:feaPartial}$ holds because the feature maps at the $(l+1)^{th}$ layer depend on those at the $l^{th}$ layer. As the partial derivative of the loss function with respect to the output feature can be simply calculated according to the loss function definition, the gradient on each image can be calculated recursively. Simple summation of the image gradients produces the overall gradient of the objective function.


We now turn to the triplet-based objective function and show that the overall gradient can also be obtained from the image-based gradients, which can be calculated separately. The difficulty lies in the impossibility of writing the objective function directly as the sum of the loss functions on the images, as in Equation $\ref{equ:imLoss}$, because it takes the following form, where $n$ is the number of triplets: 
\begin{equation}
f=\Sigma_{i=1}^{n}loss(F_W(O_i^1),F_W(O_i^2),F_W(O_i^3))
\end{equation}
However, because the loss function for each triplet is still defined on the outputs of the images in each triplet, this objective function can also be seen as follows, where $\{I'_k\}$ represents the set of all the distinct images in the triplets, i.e. $\{I'_k\}=\{O_i^1\}\bigcup\{O_i^2\}\bigcup\{O_i^3\}$ and $m$ is the number of the images in the triplets.
\begin{equation}
f=f(F_W(I'_1),F_W(I'_2),...,F_W(I'_m))
\end{equation}

As $F_W(I'_k)$ is some function of the feature map $X_k^l$ at the $l^{th}$ layer, the objective function can also be seen as follows:
\begin{equation}
f=f(X_1^l,X_2^l,...,X_m^l)
\end{equation}

Then the derivative rule gives us the following equations with $X_k^l$ depending on $W^l$ and $X_k^{l+1}$ depending on $X_k^l$.
\begin{equation}
\frac{\partial f}{\partial W^l}=\Sigma_{k=1}^{m}\frac{\partial f}{\partial X_k^l}\frac{\partial X_k^l}{\partial W^l}
\label{equ:paramPartialTriplet}
\end{equation}

\begin{equation}
\frac{\partial f}{\partial X_k^l}=\frac{\partial f}{\partial X_k^{l+1}}\frac{\partial X_k^{l+1}}{\partial X_k^l}
\label{equ:feaPartialTriplet}
\end{equation}

The first equation shows the gradient of the loss function with respect to the network parameters takes image-based form (summation over images) and tells us how to get this gradient given $\frac{\partial f}{\partial X_k^l}$ for all $k$. Actually, $\frac{\partial f}{\partial W^l}$ can be obtained by $\Sigma \alpha_k \frac{\partial f}{\partial X_k^l}$ with $\alpha_k=\frac{\partial X_k^l}{\partial W^l}$ whose computation only relies on image $I'_k$. If we get  $\frac{\partial f}{\partial W^l}$ for all the layers, then we get the overall gradient of the triplet-based loss function, i.e. $\Delta W=\frac{\partial f}{\partial W}$.
 
The second equation tells us how to get the partial derivative of the loss function with respect to the feature map of each image $I'_k$ at the $l^{th}$ layer, i.e. $\frac{\partial f}{\partial X_k^l}$ recursively. More precisely, if we have known the partial derivative with respect to the feature maps of the upper layer, say the $(l+1)^{th}$ layer, then the derivative with respect to this layer can be derived by simply multiplying a matrix $\frac{\partial X_k^{l+1}}{\partial X_k^l}$ which can also be calculated for each image $I'_k$ separately.

So if we get the partial derivative of the loss function with respect to the output (feature map of the top layer) of each image, i.e. $\frac{\partial f}{\partial F_W(I'_k)}$, we can get the gradient $\frac{\partial f}{\partial W}$ by applying Equation $\ref{equ:feaPartialTriplet}$ and Equation $\ref{equ:paramPartialTriplet}$ recursively (standard backward propagation). Luckily, the derivative with respect to the output of each image can be easily obtained as follows since it is defined analytically on $\{F_W(I'_k)\}$.
\begin{small}
\begin{equation}
\label{equ:outputDerive}
\frac{\partial f}{\partial F_W(I'_k)}=\Sigma_{i=1}^{n}\frac{\partial \max\{||F_W(O_i^1)-F_W(O_i^2)||^2-||F_W(O_i^1)-F_W(O_i^3)||^2,C\}}{\partial F_W(I'_k)}
\end{equation}
\end{small}

Algorithm $\ref{alg:partialDerivative}$ provides the details of calculating $\frac{\partial f}{\partial F_W(I'_k)}$. As the algorithm shows, we need to collect the derivative from each triplet. If the triplet contains the target image $I'_k$ and the distance $d(W,O_i)$ is greater than the constant $C$ (implementing the max operation in equation $\ref{equ:objFunc}$), then this triplet contributes its derivative with respect to $F_W(I'_k)$. The form of this derivative depends on the position where the image $I'_k$ appears in the triplet $O_i$ as listed in the algorithm. Otherwise, this triplet will be simply passed. With this image-based gradient calculation method, the whole training process is given in Algorithm $\ref{alg:gradDescentByImage}$. It is not hard to see that our optimized learning algorithm is very similar to the traditional neural network algorithm except that calculating the partial derivative with respect to the output of one image for the triplet-based loss function relies on the outputs of other images while the traditional loss function does not. This optimized learning algorithm has two obvious merits:

1. We can apply a recent deep learning implementation framework such as Caffe \cite{Jia13caffe} simply by modifying the loss layer.

2. The number of network propagation executions can be reduced to the number of distinct images in the triplets, a crucial advantage for large scale datasets.
\begin{small}
\begin{algorithm}[htb]        
\caption{ Image-based gradient descent algorithm }            
\label{alg:gradDescentByImage}                
\begin{algorithmic}[1]               
\REQUIRE ~~\\                         
   Training triplets $\{O_{i}\}$;
\ENSURE ~~\\                          
	The network parameters $W$;
\STATE Collect all the distinct images $\{I'_k\}$ in $\{O_i\}$;
\WHILE    {$t<T$}
\STATE $t\leftarrow t+1$;
\STATE $\frac{\partial f}{\partial W}=0$;
\STATE Calculate the outputs for each image $I'_k$ by forward propagation;
   \FORALL {$I'_k$} 
   	\STATE Calculate $\frac{\partial f}{\partial F_W(I'_k)}$  for image $I'_k$ according to Algorithm $\ref{alg:partialDerivative}$;

\STATE Calculate $\frac{\partial f}{\partial W}(I'_k)$ using back propagation;
\STATE Increment the partial derivative: $\frac{\partial f}{\partial W}$+=$\frac{\partial f}{\partial W}(I'_k)$;

   \ENDFOR
\STATE $W^t=W^{t-1}-\lambda_t \frac {\partial f}{\partial W}$;
\ENDWHILE          
\end{algorithmic}
\end{algorithm}
\end{small}

\begin{small}
\begin{algorithm}[htb]        
\caption{ Partial derivative with respect to the output of image $I'_k$}            
\label{alg:partialDerivative}                
\begin{algorithmic}[1]               
\REQUIRE ~~\\                         
   Training triplets $\{O_{i}\}$, image $I'_k$;
\ENSURE ~~\\                          
	The partial derivative: $\frac{\partial f}{\partial F_W(I'_k)}$;
\STATE $\frac{\partial f}{\partial F_W(I'_k)}=0$;
   \FORALL {$O_{i}=<O_{i}^1, O_{i}^2, O_{i}^3>$} 
	\IF {$d(W,O_i)>C$}
        \IF {$I'_k$=$O_{i}^1$}
            \STATE $\frac{\partial f}{\partial F_W(I'_k)}+=2(F_W(O_{i}^3)-F_W(O_{i}^2))$;
        \ELSIF {$I'_k$=$O_{i}^2$}
           \STATE $\frac{\partial f}{\partial F_W(I'_k)}-=2(F_W(O_{i}^1)-F_W(O_{i}^2))$;
        \ELSIF {$I'_k$=$O_{i}^3$}
           \STATE $\frac{\partial f}{\partial F_W(I'_k)}+=2(F_W(O_{i}^1)-F_W(O_{i}^3))$;
        \ENDIF
	\ENDIF
   \ENDFOR
\end{algorithmic}
\end{algorithm}
\end{small}

\section{Batch Learning and Triplet Generation}
Suppose that we have a labelled dataset with $M$ classes (persons) and that each class has $N$ images. The number of possible triplets would be $M(M - 1)N^2(N - 1)$. It would be impossible to load all of these triplets into the memory to train the network even for a moderate dataset. It is thus necessary to train the network using the batch mode, which allows it to be trained iteratively. In each iteration, only a small part of triplets are selected from all the possible triplets, and these triplets are used to calculate the gradient and then to update the network parameters. There are several ways to select triplets from the full population of triplets. The simplest method is to select them randomly. However, in random selection, the distinct image size is approximately three times of the selected triplet size because each triplet contains three images, and the likelihood of two triplets sharing the same image is very low. This triplet generation approach is very inefficient because only a few distance constraints are placed on the selected images in each iteration.  Instead, according to our optimized gradient derivation, we know that the number of network propagations depends on the number of images contained in the triplets. So we should produce more triplets to train the model with the same number of images in each iteration. This leads to our following triplet generation scheme. In each iteration, we select a fixed number of classes (persons), and for each image in each class, we randomly construct a large number of triplets, in which the matched references are randomly selected from the same class and the mismatched references are randomly selected from the remaining selected classes. This policy ensures large amounts of distance constraints are posed on a small number of images, which can be loaded into the limited memory in each iteration. And with the increasing number of iterations are executed, the sampled triplets still can cover all the possible triplet pattern, ensuring the model to converge to a local minimum. 

As a comparison, suppose the memory can only load 300 images (a typical case for 2G GPU memory device). Then in the random triplet generation scheme, only about 100 triplets can be applied to train the model in one iteration. However, our proposed scheme can use thousands of triplets to train the model without obvious computation load increase. Algorithm $\ref{alg:batchTraining}$ gives the complete batch training process. As described in the ablation study section, our proposed triplet generation scheme shows obvious advantages both in convergence time and matching rate.

\begin{small}
\begin{algorithm}[htb]        
\caption{Learning deep features from relative distance comparison in the batch mode}            
\label{alg:batchTraining}                
\begin{algorithmic}[1]               
\REQUIRE ~~\\                         
   Labelled training images $\{I_i\}$;
\ENSURE ~~\\                          
	Network Parameters $W$;
\WHILE    {$t<T$}
\STATE $t\leftarrow t+1$;
\STATE Randomly select a subset of classes (persons) from the training set;
\STATE Collect images from the selected  classes: $\{I'_k\}$ ;
\STATE Construct a set of triplets from the selected classes;
\STATE $\Delta W=0$;
  \FORALL {$I'_k$}
\STATE Run forward propagation for $I'_k$
\STATE Calculate the partial derivative of the loss function with respect to $F_W(I'_k)$ according to Algorithm $\ref{alg:partialDerivative}$;
\STATE Run the standard backward propagation for $I'_k$;
\STATE Accumulate the gradient: $\Delta W+=\Delta W(I'_k)$;
\ENDFOR
\STATE $W^t=W^{t-1}-\lambda_t \Delta W$;
\ENDWHILE          

\end{algorithmic}
\end{algorithm}
\end{small}

\section{Experiments}
\subsection{Datasets and Evaluation Protocol}
We used two well-known and challenging datasets, i.e., iLIDS \cite{iLIDS} and VIPeR  \cite{gray2008viewpoint}, for our experiments. Both datasets contain a set of persons, each of whom has several images captured by different cameras. All the images were resized to 250 $\times$ 100 pixels to train our network.

\textbf{iLIDS dataset} The iLIDS dataset \cite{iLIDS} was constructed from video images captured in a busy airport arrival hall. It features 119 pedestrians, with 479 images normalized to 128 $\times$ 64 pixels. The images come from non-overlapping cameras, and were subject to quite large illumination changes and occlusions. On average, there are four images of each individual pedestrian.

\textbf{VIPeR dataset} The VIPeR dataset \cite{gray2008viewpoint} contains two views of 632 pedestrians. The pair of images of each pedestrian was captured by different cameras under different viewpoint, pose and light conditions. It is the most challenging dataset in the person re-identification arena owing to the huge variance and discrepancy. 

\begin{figure}[!ht]
\begin{center}
\includegraphics [width=3.4in]{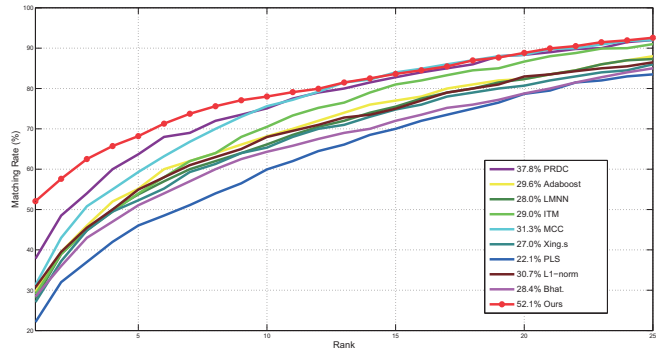}
\caption{Performance comparison using CMC curves on i-LIDS dataset.}
\label{fig:cmc-ilids}
\end{center}
\end{figure}

\textbf{Evaluation Protocol} We adopted the widely used cumulative match curve (CMC) approach \cite{gray2007evaluating} for quantitive evaluation. We randomly selected about half of the persons for training (69 for iLIDS and 316 for VIPeR), with the remainder used for testing. To obtain the CMC values, we divided the testing set into a gallery set and a probe set, with no overlap between them. The gallery set comprised one image for each person. For each image in the probe set, we returned the $n$ nearest images in the gallery set using the $L2$ distance with the features produced by the trained network. If the returned list contained an image featuring the same person as that in the query image, this query was considered as success of rank $n$. We repeated the procedure 10 times, and used the average rate as the metric.

\begin{figure}[!ht]
\begin{center}
\includegraphics [width=3.4 in]{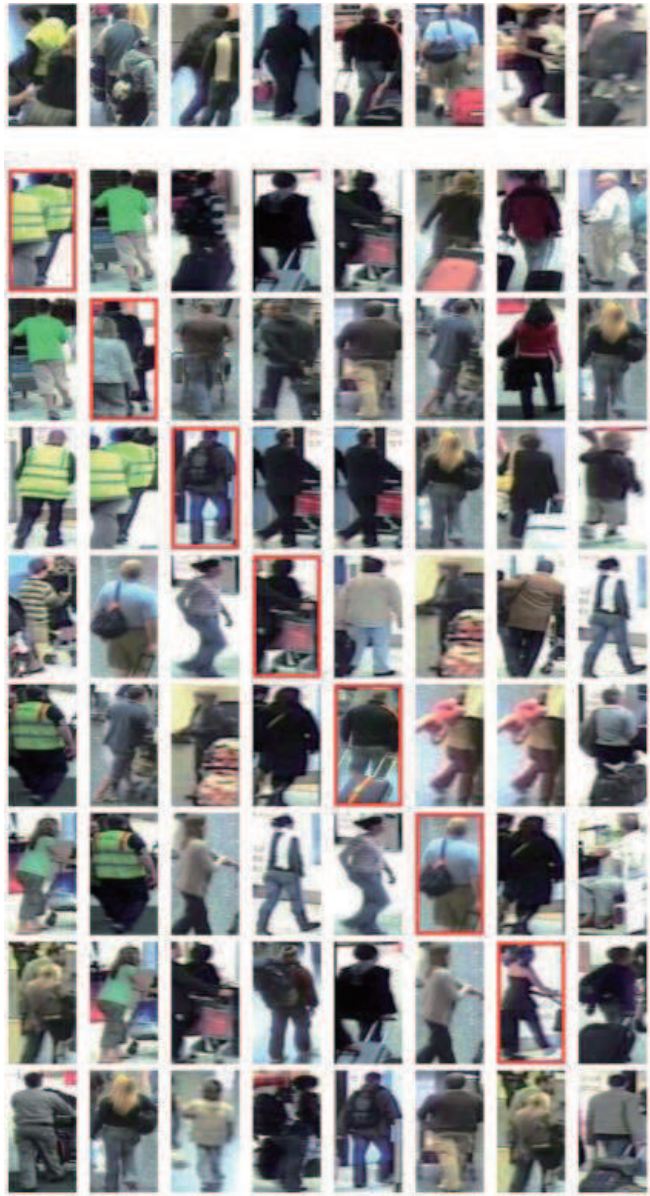}
\caption{Search examples on iLIDS dataset. Each column represents a ranking result with the top image being the query and the rest images being the returned list. The image with the red bounding box is the matched one.}
\label{fig:search-ilids}
\end{center}
\end{figure}

\begin{table}[tbp]
\small
\centering 
\begin{tabular}{lcccccc  }
\hline
Method &Top1 &Top5 &Top10 &Top15 &Top20 &Top30\\ \hline \vspace{0.1mm}  
Ours &$\textbf{52.1}$ &$\textbf{68.2}$ &$\textbf{78.0}$ &$\textbf{83.6}$ &$\textbf{88.8}$ &$\textbf{95.0}$\\ \vspace{0.1mm}
Adaboost &29.6 &55.2 &68.1 &77.0 &82.4 &92.1\\  \vspace{0.1mm}
LMNN &28.0 &53.8 &66.1 &75.5 &82.3 &91.0\\ \vspace{0.1mm}
ITML &29.0 &54.0 &70.5 &81.0 &86.7 &95.0\\  \vspace{0.1mm}
MCC &31.3 &59.3 &75.6 &84.0 &88.3 &95.0\\  \vspace{0.1mm}
Xing's &27.0 &52.3 &63.4 &74.8 &80.7 &93.0\\  \vspace{0.1mm}
PLS &22.1 &46.0 &60.0 &70.0 &78.7 &87.5\\  \vspace{0.1mm}
L1-norm &30.7 &55.0 &68.0 &75.0 &83.0 &90.0\\  \vspace{0.1mm}
Bhat. &28.4 &51.1 &64.3 &72.0 &78.8 &89.0\\  \vspace{0.1mm}
PRDC &37.8 &63.7 &75.1 &82.8 &88.4 &95.0\\ \hline

\end{tabular}
\caption{Performance of different models on i-LIDS dataset. }
\label{table:perf-ilids}
\end{table}

\begin{figure}[!ht]
\begin{center}
\includegraphics [width=3.4 in]{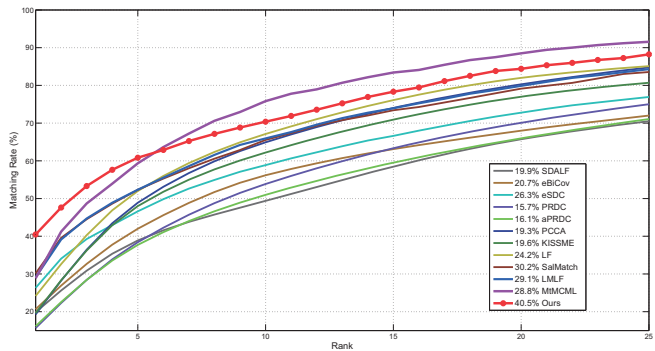}
\caption{Performance comparison using CMC curves on VIPeR dataset.}
\label{fig:cmc-viper}
\end{center}
\end{figure}

\begin{figure}[!ht]
\begin{center}
\includegraphics [width=3.4 in]{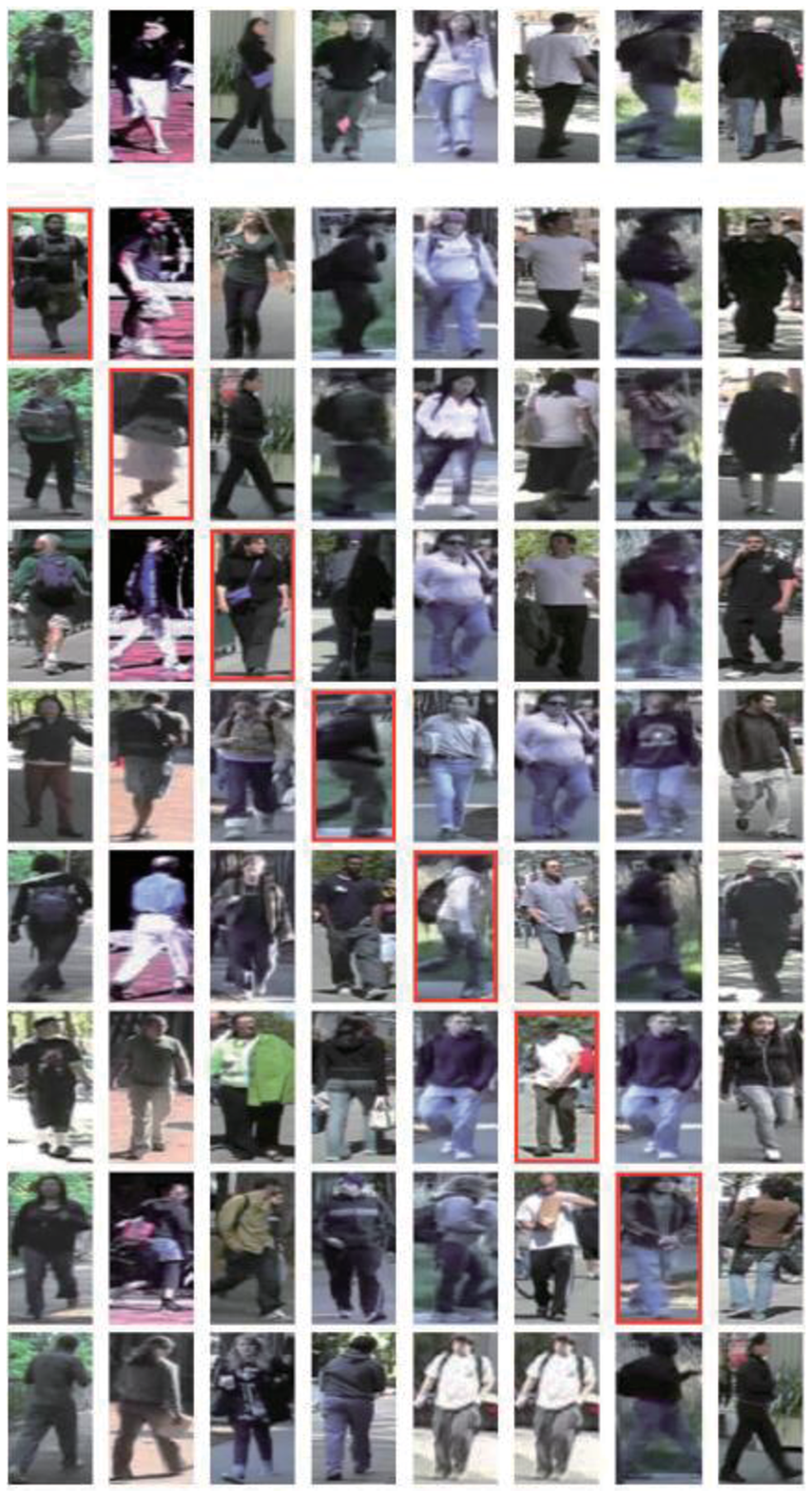}
\caption{Search examples on VIPeR dataset. Each column represents a ranking result with the top image being the query and the rest images being the returned list. The image with the red bounding box is the matched one.}
\label{fig:search-viper}
\end{center}
\end{figure}

\subsection{Performance Comparison}

\textbf{Training Setting} The weights of the filters  and the full connection parameters were initialized from two zero-mean Gaussian distributions with standard deviation 0.01 and 0.001 respectively. The bias terms were set with the constant 0. We generated the triplets as follows. In each iteration, we selected 40 persons and generate 80 triplets for each person. When there were less than 10 triplets whose distance constraints could not be satisfied, i.e. the distance between the matched pair is larger than the distance between the mismatched pair, the learning process was taken as converged. 

\textbf{Comparison on iLIDS dataset} Using the iLIDS dataset, we compared our method with PRDC \cite{zheng2011person} and other metric learning methods (i.e. Adaboost\cite{gray2008viewpoint}, Xing's \cite{xing2002distance}, LMNN \cite{weinberger2005distance}, ITML \cite{davis2007information}, PLS \cite{schwartz2009learning}, Bhat. \cite{gray2008viewpoint}, L1-norm \cite{wang2007shape} and MCC \cite{globerson2005metric}). The features were an ensemble of color histograms and texture histograms, as described in \cite{zheng2011person}. We used 69 persons for training and the rest for testing (the same setting as used by the compared methods). Figure $\ref{fig:cmc-ilids}$ shows the curves of the various models, and Table $\ref{table:perf-ilids}$ shows the top 1 , top 5, top 10, top 15, top 20 and top 30 performance. Our method achieved rank-1 accuracy 52.1\%,  which clearly outperformed the other methods. Figure $\ref{fig:search-ilids}$ shows several query examples for the iLIDS dataset. In this figure, each column represents a ranking result with the top image being the query image. The matched one in the returned list is marked by a red bounding box.

\textbf{Comparison on VIPeR dataset} Using the VIPeR dataset, we compared our method with such state-of-the-art methods as MtMCML \cite{ma2014person}, LMLF \cite{zhao2013learning}, SDALF \cite{farenzena2010person}, eBiCov \cite{ma2012bicov}, eSDC \cite{zhao2013unsupervised}, PRDC \cite{zheng2011person}, aPRDC \cite{liu2012person}, PCCA \cite{mignon2012pcca}, KISSME \cite{kostinger2012large}, LF \cite{pedagadi2013local} and SalMatch \cite{zhao2013person}. Half of the persons were used for training, and the rest for testing (the same setting as used by the compared methods). Figure $\ref{fig:cmc-viper}$ presents the CMC curves of the various models, and Table $\ref{table:per-viper}$ presents the top 1 , top 5, top 10, top 15, top 20 and top 30 ranking results. Our method achieved rank-1 accuracy 40.5\%  that clearly outperformed most available benchmarking methods. Figure $\ref{fig:search-viper}$ shows some query examples for the VIPeR dataset. Each column represents a ranking result with the top image being the query image and the rest being the result list. The matched one in the returned list is highlighted by a red bounding box. This figure shows the difficulty of this dataset. Actually, in the failed examples (rank 1 image does contain the same person as the query), the images ranked higher than the matched one often look more closer to the query image as in columns 2-7.

\begin{table}[tbp]
\small
\centering 
\begin{tabular}{lcccccc  }
\hline
Method &Top1 &Top5 &Top10 &Top15 &Top20 &Top30\\ \hline \vspace{0.1mm}  
Ours &$\textbf{40.5}$ &$\textbf{60.8}$ &70.4 &78.3 &84.4 &90.9\\ \vspace{0.1mm}
MtMCML &28.8 &59.3 &$\textbf{75.8}$ &$\textbf{83.4}$ &$\textbf{88.5}$ &$\textbf{93.5}$\\ \vspace{0.1mm} 
SDALF &19.9 &38.4 &49.4 &58.5 &66.0 &74.4\\  \vspace{0.1mm}
eBiCov &20.7 &42.0 &56.2 &63.3 &68.0 &76.0\\ \vspace{0.1mm}
eSDC &26.3 &46.4 &58.6 &66.6 &72.8 &80.5\\  \vspace{0.1mm}
PRDC &15.7 &38.4 &53.9 &63.3 &70.1 &78.5\\  \vspace{0.1mm}
aPRDC &16.1 &37.7 &51.0 &59.5 &66.0 &75.0\\  \vspace{0.1mm}
PCCA &19.3 &48.9 &64.9 &73.9 &80.3 &87.2\\  \vspace{0.1mm}
KISSME &19.6 &48.0 &62.2 &70.9 &77.0 &83.7\\  \vspace{0.1mm}
LF &24.2 &52.3 &67.1 &76.1 &82.2 &87.9\\  \vspace{0.1mm}
SalMatch &30.2 &52.3 &66.0 &73.4 &79.2 &86.0\\  \vspace{0.1mm}
LMLF &29.1 &52.3 &66.0 &73.9 &79.9 &87.9\\  \hline

\end{tabular}
\caption{Performance of different models on VIPeR dataset. }
\label{table:per-viper}
\end{table}

\subsection{Ablation Studies of Learning}

In this section, we explore the learning details on the VIPeR dataset, as it is more challenging and contains more images. 


\textbf{Data Augmentation} Data augmentation is an important mechanism for alleviating the over-fitting problem. In our implementation, we crop a center region 230 $\times$ 80 in size with a small random perturbation for each image to augment the training data. Such augmentation is critical to the performance, particularly when the training dataset is small. In our experiment, the performance declined by 33 percent without it.

\textbf{Normalization} Normalization is a common approach in CNN networks \cite{krizhevsky2012imagenet}, but these networks normalize the feature map over different channels. In our model, the output feature is normalized to 1 under the $L2$ norm.  Without this normalization, the top 1 performance drops by 25 percent.  Normalization also helps to reduce the convergence time. In our experiment, the learning process roughly converged in four 4,000 iterations with normalization and in roughly 7,000 without it.

\begin{figure}[ht]
\begin{center}
\includegraphics [width= 3.4 in]{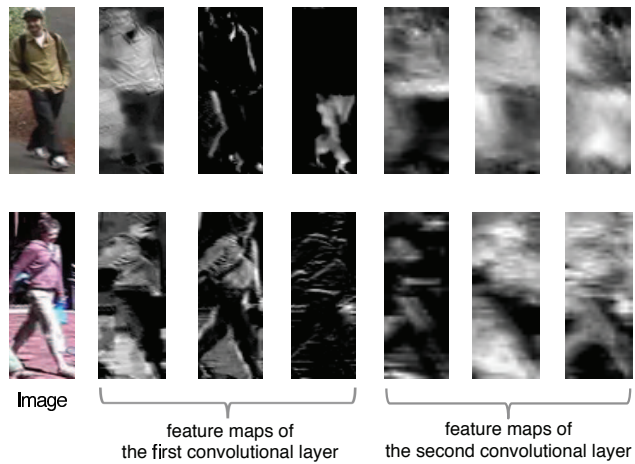}
\caption{Visualization of feature maps generated by our approach.}
\label{fig:vis_features}
\end{center}
\end{figure}

\textbf{Triplet Generation} The triplet generation scheme affects the convergence time and matching rate, as pointed out in previous sections.  We compared the model's performance under two triplet generation schemes. In the first scheme, we selected 40 persons in each iteration, and constructed 80 triplets for each person using the images of those 40 persons. In the second scheme, we again selected 40 persons in each iteration, but constructed only one triplet for each person (approximating random selection). The first scheme achieved its best performance in about 4,000 iterations while the second scheme achieved its best performance (90 percent matching rate of the first scheme) in 20,000 iterations. However, the training time in each iteration for these two schemes is almost the same as we expected. 

\textbf{Implementation Detail} We implemented our model based on the Caffe framework \cite{krizhevsky2012imagenet}, with only the data layer and loss layer replaced. We trained the network on a GTX 780 GPU with 2G memory. When there were fewer than 10 triplets whose distance constraints had been violated, the model was taken as converged. Our model usually converged in less than one hour thanks to its simplified network architecture and effective triplet generation scheme.

\textbf{Feature map visualization} In addition, we visualize the intermediate features generated by our model to validate the effectiveness of representation learning. Figure $\ref{fig:vis_features}$ shows two examples, where we present some feature maps of the first and the second convolutional layers, respectively. As we expect, the lower layer feature maps tend to have strong responses at the edges, showing some characteristics of low level features.

\section{Conclusion}
In this paper, we present a scalable deep feature learning model for person re-identification via relative distance comparison.  In this model, we construct a CNN network that is trained by a set of triplets to produce features that can satisfy the relative distance constraints organized by that  triplet set. To cope with the cubically growing number of triplets, we present an effective triplet generation scheme and an extended network propagation algorithm to efficiently train the network iteratively. Our learning algorithm ensures the overall computation load mainly depends on the number of training images rather than the number of triplets. The results of extensive experiments demonstrate the superior performance of our model compared with the state-of-the-art methods. In future research, we plan to extend our model to more datasets and tasks. 

\section*{Acknowledgement}
This work was supported by the National Natural Science Foundation of China (No. 61173082 and No.61173081), Guangdong Science and Technology Program (No. 2012B031500006), Guangdong Natural Science Foundation (No. S2013050014548), Special Project on Integration of Industry, Education and Research of Guangdong Province (No. 2012B091000101), and Program of Guangzhou Zhujiang Star of Science and Technology (No. 2013J2200067). Corresponding authors of this work are L. Lin and H. Chao.

\section*{References}

\bibliography{mybibfile}

\end{document}